\documentclass[11pt]{article}

\usepackage{acl}

\usepackage{times}
\usepackage{latexsym}
\usepackage{hyperref}

\usepackage[T1]{fontenc}

\usepackage[utf8]{inputenc}

\usepackage{microtype}

\usepackage{booktabs}
\usepackage{multirow}
\usepackage{graphicx}
\usepackage{amsmath}
\usepackage{graphicx}
\graphicspath{./figures/}

\usepackage [english]{babel}
\usepackage [autostyle, english = british]{csquotes}
\MakeOuterQuote{"}
\usepackage{blindtext}
\usepackage{algorithm}
\usepackage{algpseudocode}

\title{Towards Interpretable Summary Evaluation via Allocation of Contextual Embeddings to Reference Text Topics}

\author{Ben Schaper \\
  Technical University of Munich \\
  \small{\texttt{ben.schaper@tum.de}} \\\And
  Christopher Lohse \\
  Trinity College Dublin \\
  \small{\texttt{lohsec@tcd.ie}} \\\AND
  Marcell Streile \\
  IBM \\
  \small{\texttt{streile@de.ibm.com}} \\\And
  Andrea Giovannini \\
  IBM \\
  \small{\texttt{agv@zurich.ibm.com}} \\\And
  Richard Osuala \\
  Universitat de Barcelona \\
  \small{\texttt{richard.osuala@ub.edu}}}

\begin{document}

\maketitle

\begin{abstract}

Despite extensive recent advances in summary generation models, evaluation of auto-generated summaries still widely relies on single-score systems insufficient for transparent assessment and in-depth qualitative analysis. Towards bridging this gap, we propose the \textbf{m}ultifaceted \textbf{i}nterpretable \textbf{s}ummary \textbf{e}valuation \textbf{m}ethod (MISEM), which is based on allocation of a summary's contextual token embeddings to semantic topics identified in the reference text. We further contribute an interpretability toolbox for automated summary evaluation and interactive visual analysis of summary scoring, topic identification, and token-topic allocation. MISEM achieves a promising .404 Pearson correlation with human judgment on the TAC'08 dataset. Our code and toolbox are available at  \href{https://github.com/IBM/misem}{https://github.com/IBM/misem}%
\end{abstract}

\section{Introduction}\label{sec:intro}

Auto-generated text summaries are becoming an increasingly mature, useful, and time-saving tool in research and industry with multiple applications such as email summary generation, summarizing research papers, and simplifying knowledge management and knowledge transfer in companies \cite{El-Kassas2021AutomaticSurvey}. It is crucial for production-ready applications that auto-generated summaries do not omit critical information.

In this regard, currently used summary evaluation metrics have many known limitations, which are becoming even more apparent as natural language generator (NLG) models evolve (e.g., better paraphrasing) \cite{Gehrmann2022}. Therefore, the generated texts are becoming less assessable based on surface-level (e.g., n-gram overlap based) methods of older evaluation metrics \cite{Gehrmann2022}. Recent advances in summary evaluation leverage semantic similarity and often compute a single numerical summary evaluation score \citep{Zheng2020SentenceSummarization, Xenouleas2020SUM-QE:Model, gao2020supert}. However, as noted by \citet{Gehrmann2022}, a single numerical score alone is likely too narrow to reliably indicate the quality of a text summary.

Thus, there is a clear need for an NLG evaluation metric that provides a multifaceted and interpretable quality measure. Moreover, such a measure is needed as quality gate in industrial settings that NLG-generated summaries need to pass before being displayed to end-users.

\begin{figure}[t]
    \centering
    \includegraphics[width=0.39\textwidth]{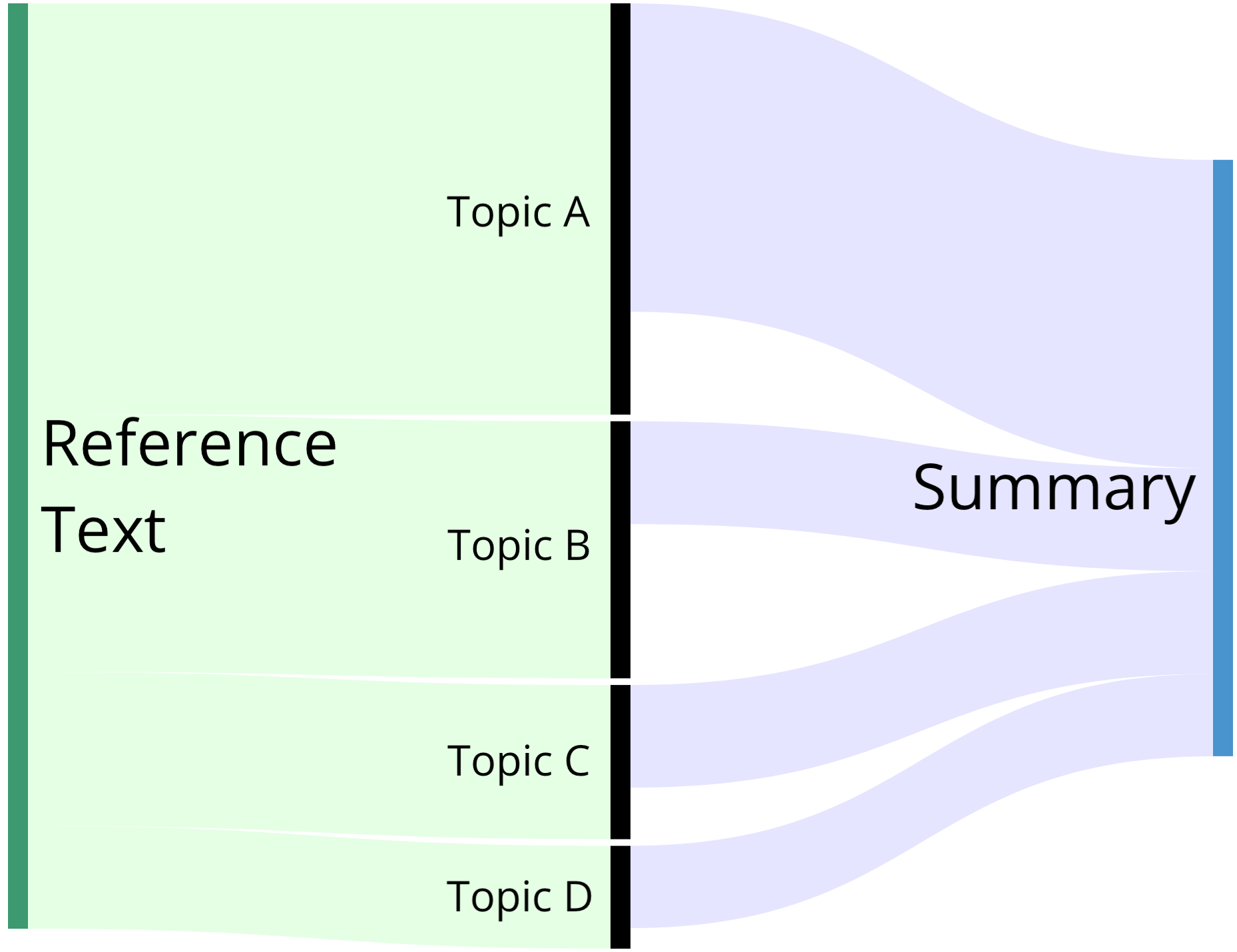}
    \caption[Sankey Diagram]{Sankey diagram visualization of the MISEM score methodology, which evaluates how well a summary reflects the topics identified in its reference text.
    }
    \label{fig:sankey}
\end{figure}
\paragraph{Contributions} 
Providing a multifaceted view of quality beyond a single score, our contributions are two-fold:

\begin{itemize}
    \item \textbf{Evaluation Method:}
    We propose an interpretable summary evaluation method that identifies semantic topics present in the reference text, assigns summary text tokens to these topics, and measures the summary's semantic coverage of each of these topics. 
    \item \textbf{Interpretability Toolbox:} %
    We provide an interactive interpretability toolbox that allows users to evaluate their summaries, adjust hyperparameters, explore topic-wise semantic overlap with reference texts, and detect missing parts of the summary.
\end{itemize}

    \label{fig:method}

\section{Related Work}

\paragraph{Evaluation based on Gold Standard} \label{par:uns}

This type of evaluation determines the relevance of a summary by comparing its overlapping words with a gold standard summary.
Such evaluation is often labor-intensive, as it requires (multiple) human written gold standard summaries. Baseline methods commonly used in gold standard evaluation are ROUGE \citep{Lin2004a} and BLEU \citep{Papineni2001}.
More recent techniques enhance the ROUGE score with WordNet \citep{ShafieiBavani2018}, word embeddings \citep{Ng2015} or use contextual embeddings \citep{Zhang2019a} in order to capture semantic similarity.
In addition to that \citet{Mrabet2020HOLMS:Models} combine lexical similarities with BERT-Embeddings \citep{Devlin2019}.

\paragraph{Annotation-based Evaluation}
Annotation-based evaluation methods require manually annotated summary ratings following predefined guidelines.
For example, the \textit{Pyramid} method \citep{Nenkova2004} works by annotating relevant topics in the source text and ranking the summaries accordingly. \citet{Bohm2020BetterReferences} use annotated texts to train a BERT-based evaluation model using rated summaries as training data.

\paragraph{Unsupervised Evaluation}

Unsupervised approaches infer a quality score of a summary based on its reference text without using a gold standard or manual annotations. Over the past few years, there have been multiple approaches to unsupervised methods for summary evaluation.
For instance, many works explore BERT-Embeddings to detect semantic similarity between summary and reference text \citep{Zheng2020SentenceSummarization, Xenouleas2020SUM-QE:Model, gao2020supert}. \citet{Zheng2020SentenceSummarization} propose a method for summary evaluation called \textit{PacSum} that combines BERT-embeddings with a directed graph in which each embedding is a node and node-wise similarity is computed based on graph positions.
In order to rate a summary, \textit{SUPERT} \citep{gao2020supert} uses contextual word embeddings and soft token alignment techniques.

\begin{figure*}[ht]
    \centering
    \includegraphics[width=0.9\textwidth]{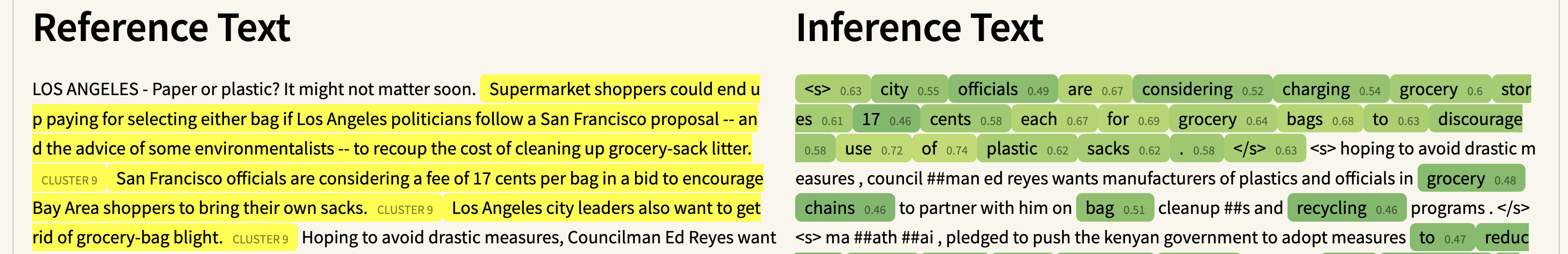}
    \caption[Annotation Tool]{Annotation Tool. By selecting one of the reference text topics (highlighted in yellow), the user can explore which summary text tokens best reflect that topic (in green). A cosine similarity threshold can be set to increase or decrease the number of highlighted summary text tokens. The summary text is referred to as 'inference' text. The example is based on text from TAC'09 \cite{Dang2008OverviewTask}.}
    \label{fig:annotation}
\end{figure*}

\section{Methodology: Summary Evaluation}

Our unsupervised evaluation method is based on the hypothesis that a generated summary can be evaluated by how much of the reference document's semantic information it preserves. To increase the granularity and interpretability of our approach, we evaluate the preservation of content per semantic topic of the reference document. The semantic topics are identified by clustering the reference document's sentence embeddings. Next, inspired by maximum mean discrepancy (MMD) \citep{gretton2012kernel}, our method measures the correspondence of the summary's contextual token embeddings to each reference text topic centroid embedding. As each token is assigned to each topic, this assignment is weighted by the normalized cosine similarity between token and topic embedding. The pseudocode implementing our method is illustrated in Algorithm \ref{alg:method}.

\begin{algorithm}[t!]
\caption{Calculation of MISEM score $m$}\label{alg:method}
\begin{algorithmic}[1]
\small
\State $R \gets encode(reference\ text)$ \Comment{Encode sentences}
\State $I \gets encode(summary\ text)$  \Comment{Encode tokens}
\State $T \gets cluster(R)$  \Comment{Cluster reference text topics}
\ForAll{$t \in T$}
    \State $w_{t} \gets \frac{|t|}{|R|}$ \Comment{Compute topic weights}
    \State $\bar{T}_t \gets \frac{1}{|t|}\sum_{i=1}^{|t|}t_{i}$\Comment{Compute topic centroids}
\EndFor
\State $C \gets \frac{\bar{T}\cdot I}{\parallel\bar{T}\parallel\parallel I \parallel}$ \Comment{Compute cosine similarity matrix}
\State $C \gets softmax(C)$ \Comment{Normalize similarity matrix}
\ForAll{$t \in T$}
\State $s_t \gets \sum_{i=1}^{n}C_{ti}$ \Comment{Compute topic scores}
\EndFor
\State $S \gets softmax(S)$ \Comment{Normalize topic scores}
\State $m = W\cdot S$ \Comment{Compute weighted final score}
\end{algorithmic}
\end{algorithm}

\paragraph{Encoding} Following \textit{SUPERT} \citep{gao2020supert},
we first split both reference text and summary text into sentences. Then, the pre-trained Sentence-BERT model \cite{Reimers2020} is
used to encode the reference text into a set of sentence embeddings $R$ and the summary text into a set of contextual token embeddings $I$. In Algorithm \ref{alg:method}, this encoding step is performed in lines 1-2.

\paragraph{Topic Clustering} Our method requires a set of reference text topics defined as clusters $T$. In our experiments, $T$ are computed using the agglomerative clustering algorithm from \citet{JMLR:v12:pedregosa11a}. Line 3 of Algorithm \ref{alg:method} contains this step. Furthermore, our method requires topic centroids $\bar{T}$, which are computed as $\bar{T}_t = \frac{1}{|t|}\sum_{i=1}^{|t|}t_{i}$, where $t$ represents one topic in $T$. Each topic centroid is calculated by taking the average of its associated reference text sentence embeddings. As the MISEM method assumes that the importance of a topic is determined to some degree by its length, a topic weight is calculated for each topic in $T$. The topic weights are calculated as $w_{t} = \frac{|t|}{|R|}$, which puts the number of sentence embeddings per topic in relation to the overall number of sentence embeddings in $R$. The topic centroid and topic weight calculations are featured in lines 4-7 of Algorithm \ref{alg:method}.

\paragraph{Summary Text Correspondence} How well a topic is reflected by the summary text tokens is calculated using cosine similarity $C=\frac{\bar{T}\cdot I}{\parallel\bar{T}\parallel\parallel I \parallel}$, which computes the similarity between the topic centroids $\bar{T}$ and the summary token embeddings $I$. Our experiments show that performance can be improved by normalizing the distribution of topic correspondence values for each token using the softmax function \cite{Bridle1989TrainingParameters}. These two steps of calculating and normalizing the similarity scores $C$ can be found in steps 8-9 of Algorithm \ref{alg:method}. For each topic $t$, the topic score is then calculated by taking the sum of its associated similarity scores $s_t = \sum_{i=1}^{n}C_{ti}$. Furthermore, we empirically find that applying the softmax normalization to $S$ improves the results. Algorithm \ref{alg:method} features the topic score calculation and subsequent normalization using the softmax function in lines 10-13.

\paragraph{Weighted Aggregation} Finally, the summary score $m$ is calculated as the sum of the topic scores $S$ scaled by the topic weights $W$ as follows: $m = W\cdot S$. In Algorithm \ref{alg:method} this final step is performed in line 14.

\paragraph{Interpretability} One concern with multiple summary scoring methods is that they only provide a single score and, as such, are limited in explaining why a particular summary was assigned a low or high score. The MISEM score addresses this problem by providing the summary text correspondence scores as part of the method. These values provide a means to examine how well a particular topic is represented by the summary, resulting in a more nuanced and transparent view of the final score. The following section introduces the interpretability toolbox, which leverages the inherent interpretability of the MISEM score to create insightful visualizations.

\section{Interpretability Toolbox}

We further contribute an interactive interpretability toolbox, which visually demonstrates the inherent interpretability of our evaluation method. As pointed out in Section \ref{sec:intro}, a single numerical score does not suffice to reliably evaluate a summary. Due to that our interpretability toolbox allows users to explore the consecutive calculations of the summary score of our method that make model behavior, various hyperparameters, and computed topic correspondence values intuitively interpretable.

\begin{figure}[ht]
    \centering
    \includegraphics[width=0.4\textwidth]{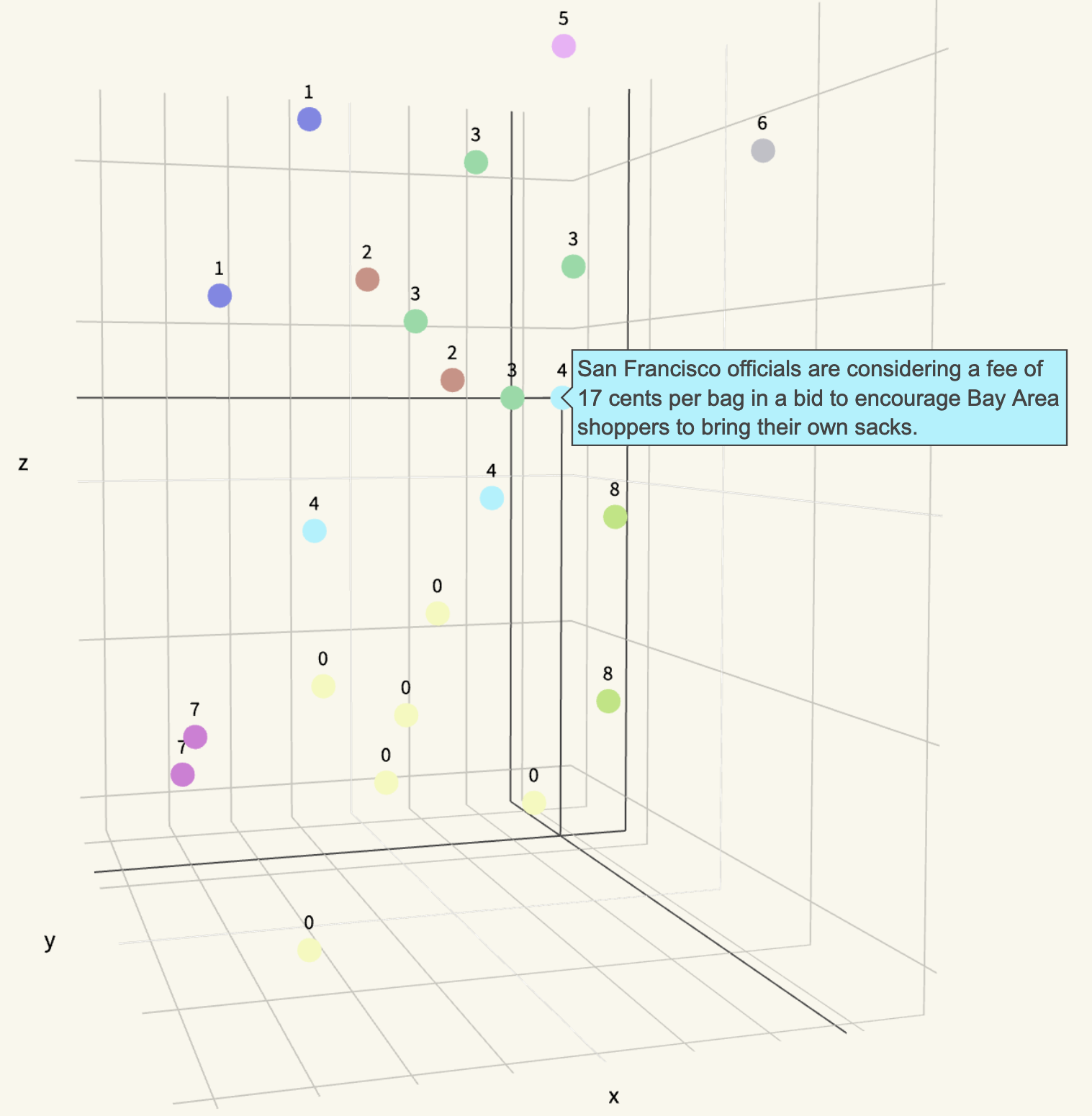}
    \caption[Scatter Plot]{Scatter Plot. Using topic-wise color-coding, the clustered sentence embeddings of the reference text are represented as spheres in a three-dimensional view. Hovering a sentence sphere shows its textual content. Example is based on text from TAC'09 \cite{Dang2008OverviewTask}.}
    \label{fig:embeddings}
\end{figure}

We propose a workflow in which users (i) initiate the summary scoring, (ii) utilize the scatter plot (see Figure \ref{fig:embeddings}) to examine the relatedness of clustered sentences within different topics, and (iii) use the annotation tool (see Figure \ref{fig:annotation}) to explore specific topics and summary text insufficiencies in-detail.

\paragraph{Scatter Plot} %
The interactive scatter plot visualizes the reference text sentence embeddings clustered into topics. The embeddings are reduced to three dimensions using t-SNE \cite{van2008visualizing}. This visualization empowers the user to explore the different topics covered in the reference text. Additionally, the user can hover individual points in the scatter plot to reveal their contents as depicted in Figure \ref{fig:embeddings}.

\paragraph{Annotation Tool} The annotation tool allows the user to selectively explore individual reference text topics (see Figure \ref{fig:annotation}). Leveraging the insights gained from the scatter plot, the annotation tool allows for further in-depth investigation. It shows the reference and summary texts displayed side-by-side to simplify the users' diagnostic process. This makes it a useful qualitative tool for users and domain experts alike to scrutinize the performance of NLG-generated summaries beyond the numerical score. Upon user selection of a topic and a similarity threshold, the summary's numerical score for that topic is visually explained showing the topic's allocated tokens with their score based on cosine similarity. %

\section{Experimental Setup and Results}

\paragraph{Datasets}
We evaluate our method using TAC'08 and TAC'09 \cite{Dang2008OverviewTask}, two multi-document summarization datasets available from the Text Analysis Conference (TAC)\footnote {\href{https://tac.nist.gov/}{https://tac.nist.gov/}}
shared tasks. Following \citet{Louis2013AutomaticallyStandard}, we only use the initial summaries (the A parts).
The TAC’08 and TAC’09 datasets include a total of 92 topics. Each topic has ten news articles, four reference summaries, and 57 (TAC’08) or 55 (TAC’09) machine-generated summaries. The average news article has 611 words and 24 sentences, while the average summary has 100 words. The Pyramid score is used as the ground-truth human rating in our experiments.

\paragraph{Related Methods}
We compare our method to two other unsupervised evaluation metrics, PacSum \cite{Zheng2020SentenceSummarization} and SUPERT \cite{gao2020supert}, which are described in Section \ref{par:uns}.

\paragraph{Note on Implementation}
Table \ref{tab:hyp_param} shows the best parameters derived from a grid search using TAC'08.
We evaluated several clustering algorithms\footnote{e.g., DBSCAN, K-Means, Fast-Clustering} and text encoding models\footnote{e.g., paraphrase-albert-small-v2, all-mpnet-base-v2} to determine the best combination.%
The best performance was achieved by combining S-BERT (\textit{all-mpnet-base-v2}) \cite{Reimers2020} contextual embeddings and agglomerative clustering from scikit-learn \cite{JMLR:v12:pedregosa11a}. Sentence splitting was done using spaCy \cite{spacy2}. The interpretability toolbox is based on streamlit\footnote{\href{https://streamlit.io/}{https://streamlit.io/}}. 
\begin{table}[ht]
\resizebox{0.5 \textwidth}{!}{%
\begin{tabular}{@{}ccc|cc@{}}
\toprule
\multicolumn{3}{c}{Agglomerative Clustering} & \multicolumn{2}{|c}{S-BERT-Model} \\ \midrule
Affinity      & Linkage       & Distance     & Model               & normalize  \\ \midrule
Cosine       & Complete      & 0.38         & all-mpnet-base-v2   & True       \\ \bottomrule
\end{tabular}%
}
\caption{Optimal Hyperparameters}
\label{tab:hyp_param}
\end{table}

\paragraph{Results} 

As summarized in Table \ref{tab:results-table}, our interpretable method shows, e.g. with a Pearson's correlation of .404 on TAC'08, promising correlation with human judgment compared to state-of-the-art methods \citep{Zheng2020SentenceSummarization, gao2020supert}. %

\begin{table}[ht]
\resizebox{0.5 \textwidth}{!}{
\begin{tabular}{@{}l|llllll@{}}
\toprule
\multicolumn{1}{c|}{\multirow{2}{*}{Method}} & \multicolumn{3}{c}{TAC’08}                                                        & \multicolumn{3}{c}{TAC’09}                                                        \\ \cmidrule(l){2-7} 
\multicolumn{1}{c|}{}                        & \multicolumn{1}{c}{$r$} & \multicolumn{1}{c}{$\rho$} & \multicolumn{1}{c}{$\tau$} & \multicolumn{1}{c}{$r$} & \multicolumn{1}{c}{$\rho$} & \multicolumn{1}{c}{$\tau$} \\ \midrule
(F1)-PacSumTopM                             & .502                    & .506                       & .381                       & .495                    & .461                       & .337                       \\
(F $\beta$)-PacSumTopM                      & .507                    & .508                       & .380                       & .500                    & .465                       & .339                       \\
MRoBERTa (SUPERT)                           & .366                    & .326                       & .235                       & .357                    & .316                       & .229                       \\
MSBERT (SUPERT)                             & .466                    & .428                       & .311                       & .436                    & .435                       & .320                       \\
MISEM (\textbf{ours})                                        & .404                    & .375                       & .270                       & .349                   & .376                      & .274                      \\ \bottomrule
\end{tabular}
}
\caption{Main results on multi-document summarization datasets. Pearson’s $r$, Spearman’s $\rho$, and Kendall’s $\tau$ correlation with human scores are reported. All scores except for MISEM were taken from their respective publications.}
\label{tab:results-table}
\end{table}

\section{Conclusion}

In this paper, we present MISEM – an unsupervised method for interpretable summary evaluation based on semantic topics extracted from the reference text. Our method aims to provide a multifaceted view of a generated summary and allows for both quantitative as well as qualitative evaluation. Our experimental results show that our method achieves a competitive correlation with the human judgement Pyramid score on the TAC’08 and TAC’09 datasets. Beyond the MISEM method, we also provide an interpretability toolbox, which allows users to %
interactively and intuitively understand the summary evaluation, explore reference text topics, and analyze summary token to topic allocation.

\paragraph{Limitations}

We suggest the application of our method as summary quality gate in conjunction with other complementary methods to overcome several limitations. The latter include that our method does not incentivize conciseness of evaluated summaries due to not penalizing their length.
Another limitation is an exclusive reliance on the semantic similarity of text embeddings. For instance, contradicting opinions (e.g., expressed via negation terms) in the summary can still yield a high semantic similarity with the respective reference text topic. %
Also, our method's focus on semantics limits its capability of capturing morphological or syntactic errors that may also affect a summary's quality. We further note that our method is constrained by the following three assumptions: (1) that topical diversity is vital for high-quality summaries, (2) that clustering of sentence embeddings provides a good representation of such topics, and (3) that computing, visualizing, and comparing the similarity between the contextual token embeddings of a summary and the centroids of clustered sentence embeddings of its reference text allows to meaningfully interpret the summary's quality.

\paragraph{Future Work}
We suggest future work to test fine-tuning the encoding model, e.g., with the goal of optimizing the creation of sentence embedding clusters. 
Moreover, promising extensions of our method and toolbox are integrated scoring of summary morphology and syntax. Also, we motivate the validation in further domains (e.g. legal and medical texts) and on multilingual data using multilingual encoders.
In addition to that, validating our approach on more recent datasets e.g. SummEval \cite{fabbri2021summeval}, or RealSumm \cite{Bhandari-2020-reevaluating} and Newsroom \cite{grusky2018newsroom} are suggested to improve the generalizability of our findings. The interpretability toolbox could potentially be enriched by the addition of an interactive Sankey diagram, as indicated in Figure \ref{fig:sankey}. %
The latter further enables users to instantly and visually detect missing, under- and overrepresented topics in the summary.

\paragraph{Closing Remarks}

Our work calls for further research on interpretable and multifaceted summary evaluation methods and shows promising potential towards closing this research gap identified by \citet{Gehrmann2022}.

\bibliography{references}
\bibliographystyle{acl_natbib}
\end{document}